\documentclass{article}

    \PassOptionsToPackage{numbers, compress}{natbib}

    \usepackage[preprint]{neurips_2024}

\usepackage[utf8]{inputenc} 
\usepackage[T1]{fontenc}    
\usepackage{hyperref}       
\usepackage{url}            
\usepackage{booktabs}       
\usepackage{amsfonts}       
\usepackage{nicefrac}       
\usepackage{microtype}      
\usepackage{xcolor}         
\usepackage{xspace}
\usepackage{graphicx}
\usepackage{subcaption}
\usepackage{multirow}
\usepackage{multicol}
\usepackage{colortbl}
\usepackage{pifont}
\usepackage{wrapfig}
\usepackage{amsmath}

\newlength\savewidth

\usepackage{pgfplots}
\usepackage{pgfplotstable}
\pgfplotsset{compat=1.8}
\usetikzlibrary{pgfplots.groupplots}
\definecolor{mediumelectricblue}{rgb}{0.01, 0.31, 0.59}
\definecolor{lightsalmon}{rgb}{1.0, 0.63, 0.48}

\makeatletter
\def\blfootnote{\xdef\@thefnmark{}\@footnotetext}
\makeatother

\title{Mini-Monkey: Alleviating the Semantic Sawtooth Effect for Lightweight MLLMs via Complementary Image Pyramid}

\author{
 Mingxin Huang \hspace{.12in} Yuliang Liu \hspace{.12in} Dingkang Liang  \hspace{.12in}  Lianwen Jin$^*$ \hspace{.12in} Xiang Bai$^*$ \\
 ylliu@hust.edu.cn
}

\begin{document}

\maketitle

\begin{abstract}
Recently, scaling images to high resolution has received much attention in multimodal large language models (MLLMs). Most existing practices adopt a sliding-window-style cropping strategy to adapt to resolution increase. Such a cropping strategy, however, can easily cut off objects and connected regions, which introduces semantic discontinuity and therefore impedes MLLMs from recognizing small or irregularly shaped objects or text, leading to a phenomenon we call the semantic sawtooth effect. This effect is particularly evident in lightweight MLLMs. To address this issue, we introduce a Complementary Image Pyramid (CIP), a simple, effective, and plug-and-play solution designed to mitigate semantic discontinuity during high-resolution image processing. In particular, CIP dynamically constructs an image pyramid to provide complementary semantic information for the cropping-based MLLMs, enabling them to richly acquire semantics at all levels. Furthermore, we introduce a Scale Compression Mechanism (SCM) to reduce the additional computational overhead by compressing the redundant visual tokens. Our experiments demonstrate that CIP can consistently enhance the performance across diverse architectures (e.g., MiniCPM-V-2, InternVL2, and LLaVA-OneVision), various model capacity (1B$\rightarrow$8B), and different usage configurations (training-free and fine-tuning). Leveraging the proposed CIP and SCM, we introduce a lightweight MLLM, Mini-Monkey, which achieves remarkable performance in both general multimodal understanding and document understanding. On the OCRBench, the 2B-version Mini-Monkey even surpasses the 8B model InternVL2-8B by $12$ score. Additionally, training Mini-Monkey is cheap, requiring only eight RTX $3090$ GPUs. The code is available at \url{https://github.com/Yuliang-Liu/Monkey}.

\end{abstract}
\let\thefootnote\relax\footnotetext{Y. Liu, D. Liang, and X. Bai are with Huazhong University of Science and Technology. M. Huang and L. Jin are with South China University of Technology. This work was done when M. Huang was visiting Huazhong University of Science and Technology. $^*$Corresponding authors.}

Recently, Large Language Models (LLMs)~\cite{zhang2022opt,brown2020gpt3,touvron2023llama,openai2023gpt4} have 
received significant attention for their robust text understanding and generation capabilities. Researchers are actively exploring ways to integrate vision encoders into LLMs 
to upgrade them to multimodal large language models (MLLMs)~\cite{li2023blip2,liu2023llava,bai2023qwen-vl}. Some approaches employ a Q-former~\cite{alayrac2022flamingo,li2023blip2}, while others~\cite{liu2024visual,wang2023cogvlm} use linear projection. Despite the promising results, they are constrained to processing low-res images, which limits their ability to 
execute detailed scene analysis.

\begin{figure*}[t!]
    \centering
    \includegraphics[width=\linewidth]{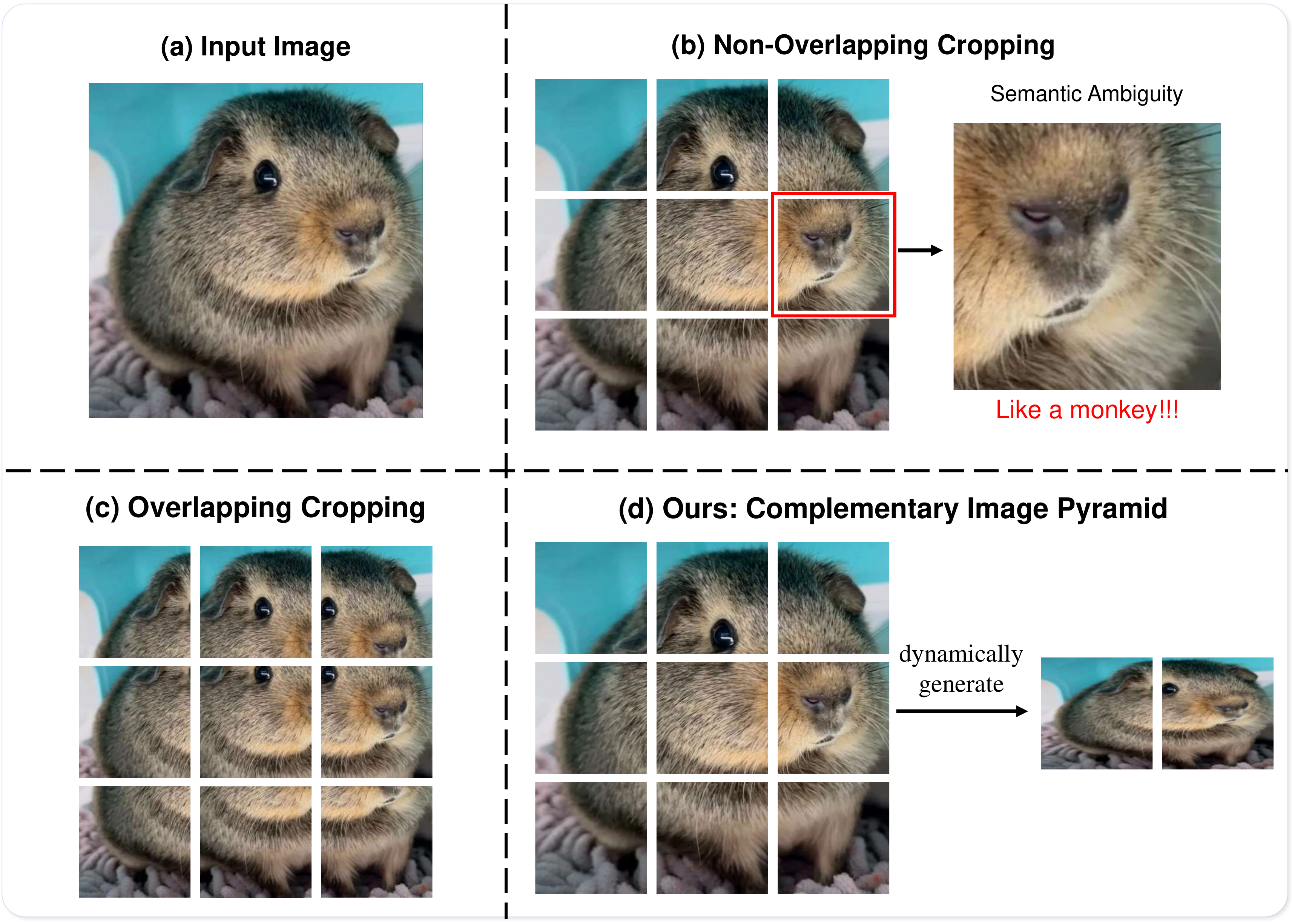}
    \caption{ 
    \textbf{Comparison of different image cropping strategies}. (a) Input image; (b) Non-overlapping cropping; (c) Overlapping cropping; (d) Ours: complementary image pyramid.}
    \label{fig:intro}
\end{figure*}

To address this limitation, 
much recent effect aims to enable MLLMs to process high-res images. One straightforward solution is to adopt a visual encoder that can 
tackle high-res images. However, developing a high-quality visual encoder demands substantial training resources~\cite{bai2023qwen-vl,chen2023pali-x}. An alternative, more resource-efficient strategy is the non-overlapping cropping~\cite{lin2023sphinx,liu2024llavanext,ye2023ureader,li2024monkey,chen2024far}, which 
splits a high-res image into a set of low-res sub-images.

While the non-overlapping cropping strategy has shown promising results, it inevitably
cuts off objects and connected regions, 
rendering difficulty for the MLLM in recognizing small or irregularly shaped objects due to semantic discontinuity, particularly in the context of document understanding. This mainly leads to two types of 
consequences: i) \textit{semantic ambiguity}: if an object or character is divided, it may be misidentified; the nose of guinea pig in Fig.~\ref{fig:intro}(b) looks
much like a monkey after cropping, for instance; 2) \textit{semantic damage}: if a word or sentence is segmented, the 
meanings of the segmented word will be 
changed completely; if the word `breakdown' 
is divided into `break' and `down', 
the segmented words will have nothing to do with the original one~\cite{liu2024textmonkey,zhang2024dockylin}. For simplicity, we call these 
phenomena 
the \textit{semantic sawtooth effect} in this paper. To alleviate this effect, a 
rather straightforward idea is to adopt 
overlapping cropping. However, this 
strategy will result in 
the processing of much duplicate information, as presented in Fig.~\ref{fig:intro}(c). This redundancy could potentially cause hallucinations in MLLMs. Moreover, it can even deteriorate 
performance  
according to our ablation studies in Sec.~\ref{sec:ab}. Additionally, the semantic sawtooth effect 
can be observed more evidently in lightweight MLLMs. Larger MLLMs with enhanced comprehension capabilities and feature extraction capabilities often can alleviate this issue to some extent. Even when the object is segmented, these models can understand the objects through their powerful feature extraction.

To alleviate the semantic sawtooth effect more explicitly, we propose a plug-and-play 
approach, termed Complementary Image Pyramid (CIP). CIP can be easily integrated into a variety of cropping-based MLLMs, allowing them to tackle high-res images with reduced semantic sawtooth effect. CIP dynamically constructs an image pyramid that provides complementary semantic features for the MLLMs, enabling it rich acquire semantics at all levels. If object semantics are lost at one scale, they can be compensated by those from another scale. Different from previous work~\cite{liu2024textmonkey,huang2024hires} that addresses this issue by modifying the architecture of the model, our approach focuses on enriching the image semantics per se. Consequently, CIP can be easily integrated into a variety of MLLMs, allowing them to tackle high-res images with reduced semantic sawtooth effect. Considering that the CIP introduces some additional computational overheads, we further propose a Scale Compression Mechanism (SCM) for use in situations with limited computational resources. The SCM is both training-free and parameter-free. It leverages the well-trained attention layers of the LLM and the multi-scale information to generate attention weights, which in turn are used to compress redundant tokens. Utilizing the proposed CIP and SCM, we introduce a lightweight MLLM, Mini-Monkey.

Our experiments demonstrate the effectiveness of the proposed method: 1) 2B-parameter Mini-Monkey outperforms the InternVL2-2B by an average of $2.4\%$ across 17 benchmarks in terms of evaluation metrics; 2) Mini-Monkey achieves a score of 806 on the OCRBench, outperforming the 8B-parameter model InternVL2-8B by $12$ score. Moreover, we observe that directly fine-tuning well-performing pre-trained MLLM does not enhance, but rather degrades its performance. In contrast, fine-tuning with CIP can facilitate the training process to improve performance. In conclusion, the contributions of this work can be summarized as follows:
\begin{itemize}
\item 
CIP: a plug-and-play complementary image pyramid designed to alleviate the semantic sawtooth effect for multimodal large language models;

\item 
Mini-Monkey: a lightweight, effective, and training-efficient multimodal large language model that integrates the complementary image pyramid and the scale compression mechanism;

\item 
Our method achieves promising results on 8 general multimodal understanding benchmarks and 9 document understanding benchmarks,  demonstrating the benefits of alleviating the semantic sawtooth effect.

\end{itemize}

\section{Related Work}

\subsection{Multimodal Large Language Models}
In recent years, Large Language Models (LLMs) have made significant progress~\cite{zhang2022opt,brown2020gpt3,touvron2023llama,openai2023gpt4}. Drawing from this advancement, many efforts have been made to integrate a vision encoder into Large Language Models for vision-language understanding. A commonly employed approach is the linear projector method\citep{liu2024visual,wang2023cogvlm}, which maps the output of the vision encoder to the same feature space as the text features of the Large Language Models. Some methods, such as Q-Former~\cite{li2023blip2}, Perceiver Resampler~\cite{alayrac2022flamingo}, or Abstractor~\cite{ye2023mplug}, introduce a set of learnable queries to facilitate this integration. Despite these notable advances, previous methods often struggle with detailed scene understanding due to limitations in resolution.

To address this issue, recent research has adopted two primary strategies: 1) The direct use of visual encoders that support high-res input. Some methods~\cite{wei2023vary,hong2024cogagent} utilize two vision encoders, one for processing high-res images and another for low-res images. Others~\cite{wang2024qwen2,liu2024oryx} leverage a single vision encoder. However, these methods require additional training data and parameters, or processing attention over high-res images significantly increases computational demands. 2) An alternative, more resource-efficient method is the cropping strategy, which divides the high-res image into multiple lower-resolution sub-images for processing. Some methods apply a fixed-size cropping scheme~\cite{li2024monkey,lin2023sphinx}. Others adopt a dynamic cropping approach~\cite{ye2023ureader,chen2024far,dubey2024llama} to keep the aspect ratio of the original image. Although the cropping strategy achieves promising results on several multimodal benchmarks, it will inevitably result in a semantic sawtooth effect: 1) If an object or character is divided, it may not be recognized; 2) If the word or sentence is segmented, the semantic damage of the segmented word will be caused. For example, the word `Breakdown' may be divided into `Break' and `down', causing semantic damage to the segmented word. This will limit the model's ability to understand the detailed scene. Although some methods~\cite{liu2024textmonkey,huang2024hires} attempt to address this issue by introducing new modules, they introduce additional parameters to the original model and require training this module from scratch. In contrast, the proposed CIP is designed to be seamlessly integrated without introducing additional parameters, offering a plug-and-play solution. A recent work~\cite{shi2024we} also attempts to utilize the multi-scale image to enhance the model's capacity. However, they employ a fixed resolution, lacking the ability to dynamically generate complementary semantics based on the resolution of input images. On the contrary, The CIP dynamically constructs an image pyramid that provides complementary semantic features after cropping for the MLLMs, enabling them to richly acquire semantics at all levels.

\subsection{Lightweight Multimodal Large Language Models}
Due to the substantial computational costs associated with multimodal large language models (MLLMs), some recent efforts have focused on developing more efficient models for rapid development and real-world applications. For instance, LLaVA-Phi~\cite{zhu2024llava} and Imp~\cite{shao2024imp} integrate a lightweight large language model with a vision encoder to develop a powerful multimodal system. MobileVLM~\cite{chu2023mobilevlm} further conserves resources by integrating a lightweight downsampling projector that reduces the number of visual tokens. Bunny~\cite{he2024bunny} advances efficiency through an effective data compression technique, which minimizes the required pretraining dataset. TinyGPT-V~\cite{yuan2023tinygpt} adopts a multi-stage training process specifically designed for lightweight multimodal models. The above models support only low-res input. To improve detailed scene understanding in lightweight MLLMs, one of the most commonly used methods is the cropping strategy. For instance, MiniCPM-V series~\cite{yao2024MiniCPM} d employs an adaptive visual encoding method to accommodate high-res images of varying aspect ratios. Similarly, InternVL2-2B~\cite{chen2024far} enhances the performance of lightweight MLLMs by adopting a dynamic high-res cropping strategy. Despite these advancements, the cropping strategy will introduce a semantic sawtooth effect, which significantly limits the performance of lightweight multimodal large language models. Larger MLLMs with enhanced comprehension capabilities can alleviate this issue to some extent, as discussed in Sec.~\ref{qualitative}.

\begin{figure*}[t!]
    \centering
    \includegraphics[width=\linewidth]{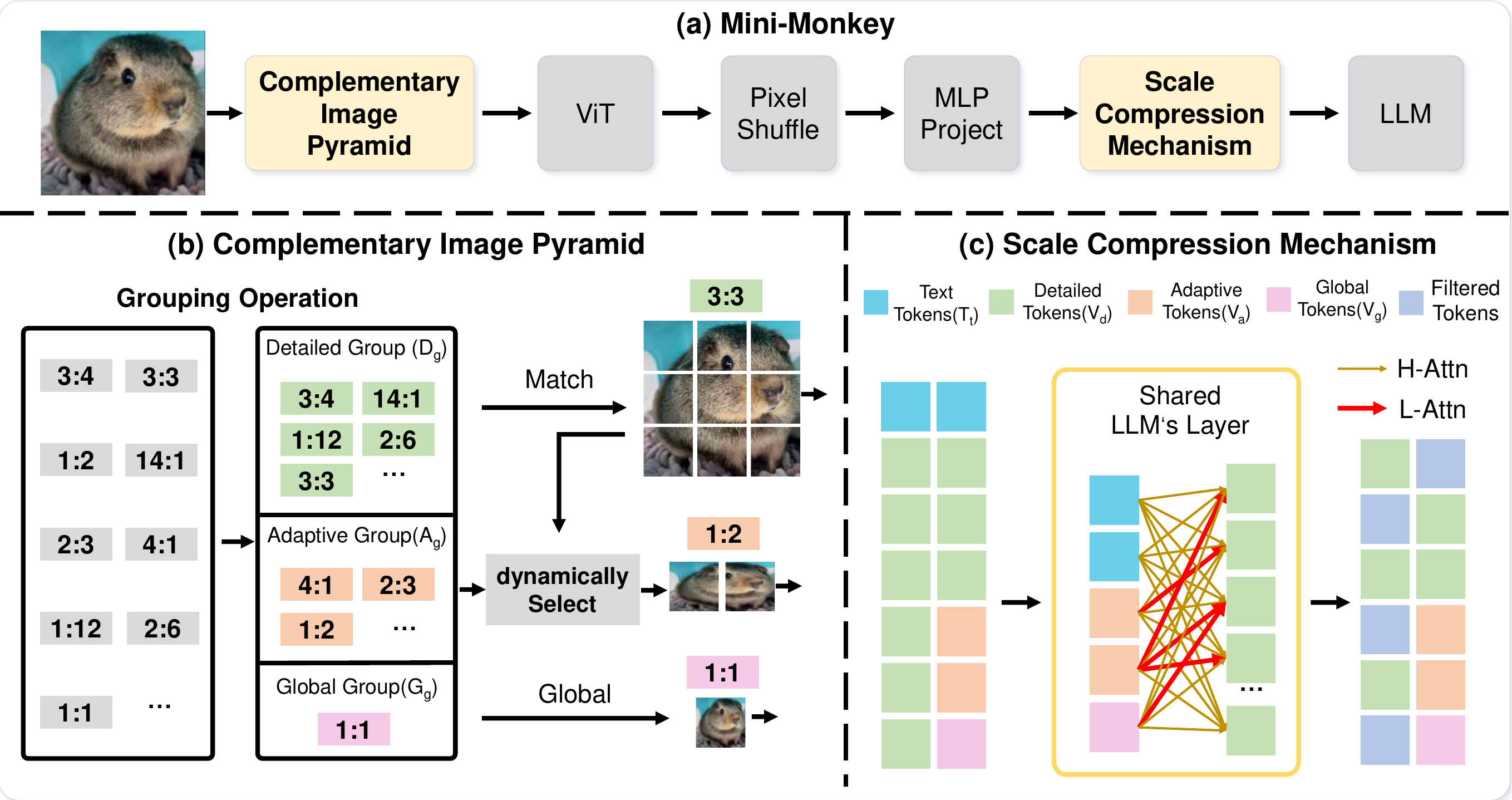}
    \caption{\textbf{Overall architecture of Mini-Monkey}.  H-Attn represents high attention weights. L-Attn represents low attention weights. The tokens with low attention weights will be filtered. The shared LLM's Layer represents using the block layer from LLM in SCM.}
    \label{fig:overvall}
\end{figure*}

\section{Mini-Monkey}
The overall architecture is illustrated in Fig.~\ref{fig:overvall}. Mini-Monkey consists of a CIP, a vision encoder, an MLP layer, a Scale Compression Mechanism, and a Large Language Model (LLM). Initially, CIP dynamically generates an image pyramid based on the resolution of input images. Then, we divide these images into a set of sub-images. These sub-images are then processed by the vision encoder and MLP layer to extract image tokens. The Scale Compression Mechanism adjusts these image tokens and forwards them to the LLM, which subsequently generates the final answers.

\subsection{Complementary Image Pyramid}
Existing cropping strategy~\cite {li2024monkey,chen2024far} directly divides the high-res images into a set of sub-images that will lead to a semantic sawtooth effect. To address this issue, we propose a plug-and-play method, termed complementary image pyramid (CIP), to promote synergy among images at varying scales to alleviate the semantic sawtooth effect. The process of CIP is shown in Fig.~\ref{fig:overvall} (b). 

\textbf{Grouping Operation.} We first generate a set of pre-defined aspect ratios. During testing, we limit the maximum number of tiles to 24.  A greater number of tiles corresponds to images with higher resolution. These aspect ratios are then categorized into three groups through a grouping operation, including a detailed group $D_g$, an adaptive group $A_g$, and a global group $G_g$. The classification is based on the following criteria: (1) Aspect ratios that result in more than 9 tiles being allocated to the detailed group, enabling the largest possible image size and thus a clearer depiction of the objects within. (2) For aspect ratios producing between 3 and 8 tiles, we classify them into the adaptive group, which is responsible for enhancing the fine details at the borders of the crops. (3) The 1:1 aspect ratio is designated to the global group, providing a low-res, comprehensive view of the whole image. The grouping operation generates three groups of different aspect ratios for generating the image.

\textbf{Dynamically Generating Images.} After the grouping operation, we will select one aspect ratio from each group to produce a total of three images. For the detailed group, we calculate the aspect ratio of the input image and then compare it with the aspect ratios within the detailed group by calculating the absolute differences. The aspect ratio that best matches the input image, denoted as $D_h, D_w$, where $D_h$ is the number of height tiles and $D_w$ is the number of width tiles, is then used to resize the image and divide the resized image into a series of sub-images. After obtaining the detailed image, the adaptive group will dynamically generate an aspect ratio based on $D_h, D_w$, ensuring that the cropping lines on the detailed group and those on the adaptive group do not overlap. To achieve this, we ensure that the aspect ratios of the adaptive and detailed groups are not integer multiples of one another. Consequently, we will dynamically remove any aspect ratios that are exact multiples of $D_h, D_w$.
This can be formulated as follows:
\begin{equation}
\forall k \in \mathbb{Z}, \, D_h \neq k \cdot A_h \quad \text{and} \quad D_w \neq k \cdot A_w
\end{equation} 
where $A_h$ and $A_w$ denote the height and width components of the aspect ratios in the adaptive group, respectively. Then, we will resize the image by selecting the ratio closest to the aspect ratio of the original image from the remaining aspect ratio. 

Due to the independent processing of sub-images in the vision encoder, previous cropping-based MLLMs have not facilitated feature interaction across different sub-images. In our method, adaptive group employ a distinct aspect ratio to partitioning windows compared to detailed group, thereby simulating cross-window feature interactions within the detailed group. The detailed image provides detailed information about the image. The adaptive image enhances the fine features at the cropping boundaries of the detailed image. The global image captures the global information of the entire image for the model. Different from previous method~\cite{liu2024textmonkey,huang2024hires}, the proposed CIP alleviates the semantic sawtooth effect from the perspective of the image, bringing several advantages: (i) it is plug-and-play, requiring no additional parameters; (ii) it seamlessly integrates with existing MLLMs that utilize cropping strategies, leading to consistent performance improvements; and (iii) it can be utilized without training and its effectiveness can be further improved through fine-tuning.

\subsection{Scale Compression Mechanism}
Although the proposed CIP significantly enhances model performance, certain scenarios may restrict the level of computational resources available. To tackle this challenge, we introduce a parameter-free token compression method called the Scale Compression Mechanism (SCM), which is used to reduce the visual tokens, as shown in Fig.~\ref{fig:overvall} (c). The detailed group provides tokens with lower information density, whereas the adaptive and global groups yield tokens that are more information-dense. Therefore, we primarily focus on compressing the tokens from the detailed group. Specifically, a well-trained LLM from MLLM can effectively select the necessary visual features based on the input question. Consequently, SCM utilizes the first and second layers of the LLM to select visual tokens without generating any additional parameters. The input visual token including $\mathbf{V_d} \in \mathbb{R}^{L_1 \times C}$, $\mathbf{V_a} \in \mathbb{R}^{L_2 \times C}$, and $\mathbf{V_g} \in \mathbb{R}^{L_3 \times C}$, and the textual token $\mathbf{T_t} \in \mathbb{R}^{T \times C}$ will be sent into an LLM's Layer. $\mathbf{V_d}$ represents the tokens from the detailed group. $\mathbf{V_a}$ represents the tokens from the adaptive group. $\mathbf{V_g}$ represents the tokens from the global group. Notable, we reuse the layer of the LLM as this LLM's Layer. The LLM's Layer will output an attention map. We use the visual token from the adaptive group, global group, and textual token to attend to the visual token from the detailed group. The calculation of the attention can be formulated as follows: 
\begin{equation}
\mathbf{Q} = {\tt cat}(\mathbf{V_a}, \mathbf{V_g}, \mathbf{T_t})\,,
\end{equation} 
\begin{equation}
\mathbf{Attn_w} = {\tt softmax}(\frac{\mathbf{Q} + {\rm PE}(\mathbf{Q}))(\mathbf{V_d} + {\rm PE}(\mathbf{V_d}))^T}{\sqrt{D}})\,.
\end{equation} 
where $\rm PE$ represents the position encoding and $D$ denotes the dimension of the LLM. $Cat()$ represents the sequence concatenation operation. After computing the attention mechanism, we average the first dimension of the attention map $\mathbf{Attn_w} \in \mathbb{R}^{(L_2+L_3+T) \times L_1}$ to obtain a weight vector $\mathbf{W_a} \in \mathbb{R}^{L_1}$. Subsequently, we select the top $K$ visual features from detailed layers based on this weight vector $\mathbf{W_a}$. These selected tokens, along with tokens from the adaptive group, global group, and textual token, are input into the LLM to generate the results. Compared to FastV~\cite{chen2024image}, SCM works in conjunction with the CIP and is more targeted by using tokens with high relative information density to compress tokens with low information density.

\section{Experiments}
\subsection{Implementation Details}
We use InternVL2-2B~\cite{chen2024far} as the Baseline to develop the Mini-Monkey. Following previous work~\cite{chen2024internvl}, we use the (448, 448) as the input resolution of InternViT. The training datasets used to train the model include DocVQA~\cite{mathew2021docvqa}, ChartQA~\cite{masry2022chartqa}, DVQA~\cite{kafle2018dvqa}, AI2D~\cite{kembhavi2016ai2d}, GeoQA+~\cite{cao2022augmented}, and LLaVA-150K (zh)~\cite{liu2024visual}. 
We use the AdamW~\cite{adamw} as the optimizer. The base learning rate is 4e-8.

\textbf{Evaluation.} Following the previous work~\cite{he2024bunny,chen2024far}, we evaluate Mini-Monkey on eleven general multimodal understanding benchmarks, including MathVista testmini~\cite{lu2023mathvista}, SEED Image~\cite{li2023seed}, RealWorldQA~\cite{xai2024grokv}, AI2D test~\cite{kembhavi2016ai2d}, POPE~\cite{li2023evaluating}, CCBench~\cite{liu2023mmbench}, MME~\cite{fu2023mme}, and HallusionBench~\cite{guan2023hallusionbench}. For document understanding, following the previous work~\cite{liu2024textmonkey}, we employ two distinct types of metrics to verify the performance of Mini-Monkey. Initially, we leverage the standard metrics provided by the benchmarks to evaluate Mini-Monkey. We utilize benchmarks such as ChartQA~\cite{masry2022chartqa}, DocVQA~\cite{mathew2021docvqa}, InfoVQA~\cite{mathew2022infographicvqa}, TextVQA~\cite{singh2019towards}, STVQA~\cite{STVQA}, FUNSD~\cite{FUNSD}, SROIE~\cite{SROIE}, POIE~\cite{kuang2023visual} and OCRBench~\cite{liu2023hidden}. We also apply the accuracy metric to verify the performance. For this metric, a response from Mini-Monkey that fully captures the ground truth is considered a true positive. Further details on this metric and the used benchmarks can be referenced in appendix~\ref{more_implementation}. 



\begin{table*}[t!]
\scriptsize
\caption{Comparison with SoTA models on 8 multimodal benchmarks.
General multimodal benchmarks encompass: MME~\cite{fu2023mme}, RealWorldQA~\cite{xai2024grokv}, AI2D test~\cite{kembhavi2016ai2d}, CCBench~\cite{liu2023mmbench}, SEED Image~\cite{li2023seed}, HallusionBench~\cite{guan2023hallusionbench}, and POPE~\cite{li2023evaluating}. 
Additionally, the math dataset includes MathVista testmini~\cite{lu2023mathvista}. The MME results we report are the sum of the perception and cognition scores. $^\mathsection$ represents the results from the OpenCompass leaderboard~\cite{2023opencompass}.
}
\centering
\setlength\tabcolsep{0.9pt}
\renewcommand{\arraystretch}{1.15}
\resizebox{0.8\linewidth}{!}{\begin{tabular}{l|c|ccccccc|c}  \toprule
                                       &   & \multicolumn{7}{c|}{General Multimodal Benchmarks}  & \multicolumn{1}{c}{Math}\\
\multirow{-2}{*}{model}  & \multirow{-2}{*}{\#param}    & MME           & RWQA        & AI2D          & CCB      & SEED           & HallB   & POPE    & MathVista \\
\hline
QWEN-VL~\cite{bai2023qwen-vl}      & 7B   & 1848.3        & 49.3$^\mathsection$           & 63$^\mathsection$                 & 65.7$^\mathsection$     & 52.5$^\mathsection$           & 29.9$^\mathsection$     & 70$^\mathsection$       &  34.9$^\mathsection$     \\
Mini-Gemini~\cite{li2024miniGemini}      & 35B   & 2141.0        &$-$           & $-$                 & $-$     & $-$           & $-$     & $-$       &  43.3     \\
LLaVA-NeXT~\cite{liu2024llavanext}      & 35B  & 2028.0        &$-$           & 74.9        & 49.2     & 75.9	       & 34.8    & \textbf{89.6}$^\mathsection$    &  46.5     \\

InternVL 1.2~\cite{chen2024internvl}               & 40B  & 2175.4        & \textbf{67.5}          & 79.0         & 59.2       & 75.6     & 47.6     & 88.0       &  47.7     \\
InternVL 1.5~\cite{chen2024far}   & 26B & \textbf{2187.8}        & 66.0          & \textbf{80.7}       & \textbf{69.8}       & \textbf{76.0} & \textbf{49.3} & 88.3 &  \textbf{53.5} \\
\hline

DeepSeek-VL~\cite{lu2024deepseek} & 1.7B & 1531.6        & 49.7$^\mathsection$         & 51.5$^\mathsection$       
 & 37.6$^\mathsection$       & 43.7$^\mathsection$ & 27.6$^\mathsection$  & 85.9$^\mathsection$  &  29.4 \\
Mini-Gemini~\cite{li2024miniGemini} & 2.2B & 1653.0        & -         & -       
 & -         & -          & -  & -  &  29.4 \\
Bunny-StableLM-2~\cite{he2024bunny}   & 2B & 1602.9        & -          & -          
 & -            & 58.8 & -  & 85.9  &  - \\ 
MiniCPM-V-2~\cite{yao2024MiniCPM}  & 2.8B & 1808.6        & 55.8$^\mathsection$          & 62.9$^\mathsection$        & 48.0$^\mathsection$             & - & 36.1$^\mathsection$  & 86.3$^\mathsection$ &  38.7 \\
InternVL 2~\cite{chen2024far}   & 2B & 1876.8        & 57.3          & 74.1       
 & 74.7             & 70.9$^\mathsection$ & 37.9  & 85.2$^\mathsection$ &  46.3 \\
\rowcolor{gray!15}
Mini-Monkey (ours)   & 2B & \textbf{1884.2}   & \textbf{57.9}     & \textbf{74.8}  & \textbf{75.5}      & \textbf{71.3} & \textbf{38.8}  & \textbf{88.0} &  \textbf{47.3} \\
\bottomrule
\end{tabular}}
\label{tab:generate_results}
\end{table*}

\begin{table*}[!t]
    \small
    \caption{Comparison to state-of-the-art MLLMs on OCR-related Tasks. Mini-Monkey achieves the best results among the 2B-parameter MLLMs. $^\mathsection$ represents the results from the OpenCompass leaderboard~\cite{2023opencompass}.}
    \centering
    \setlength{\tabcolsep}{1.5mm}{
    \resizebox{1\linewidth}{!}{\begin{tabular}{@{}l|l|lllll@{}}
    \toprule
        Model & Model Size & DocVQA$^{Test}$ & ChartQA$^{Test}$ & InfoVQA$^{Test}$ & TextVQA$^{Val}$ & OCRBench  \\ \midrule
        
        TextMonkey~\cite{liu2024textmonkey} & 9B & 73.0 & 66.9 & 28.6 & 65.6 & 558  \\ 
        TextHawk~\cite{yu2024texthawk} & 7B & 76.4 & 66.6 & 50.6 & --- & ---  \\ 
        DocKylin~\cite{zhang2024dockylin} & 7B & 77.3  & 46.6  & 66.8 & --- & ---  \\ 
        
        HiRes-LLaVA~\cite{huang2024hires} & 7B & 74.7 & 61.5 & 48.0 & 65.4 & ---  \\ 
        LLaVA-UHD~\cite{xu2024llava-uhd} & 13B  & --- & --- & --- & 67.7 & ---  \\ 
        CogAgent~\cite{hong2024cogagent} & 17B  & 81.6 & 68.4 & 44.5 & 76.1 & 590  \\ 
        UReader~\cite{ye2023ureader}  & 7B  & 65.4 & 59.3 & 42.2 & 57.6 & --- \\ 
        DocOwl 1.5~\cite{hu2024mplug} & 8B  & 82.2 & 70.2 & 50.7 & 68.6 & ---  \\ 
        HRVDA~\cite{liu2024hrvda}  & 7B  & 72.1 & 67.6 & 43.5 & --- & ---  \\ 
        TextSquare~\cite{tang2024textsquare} & 7B  & 84.3 & 79.4 & 51.5 & 66.8 & 622  \\ 
        IXC2-4KHD~\cite{dong2024internlm} & 8B  & 90.0 & 81.0 & 68.6 &77.2 & 675\\ 
        InternVL 1.5~\cite{chen2024far}  & 26B & 90.9 & \textbf{83.8} & 72.5 & \textbf{80.6} & 724 \\ 
        InternVL 2~\cite{chen2024far} & 8B & \textbf{91.6} & 83.3 & \textbf{74.8} & 77.4 & \textbf{794} \\ 
        GLM4-V~\cite{glm2024chatglm} & 9B & - & - & - & - & 786 \\ 
        \midrule
        Vary-toy~\cite{wei2024small} & 1.8B & 65.6 &  59.1 & - & - & - \\ 
        MiniCPM-V 2.0~\cite{yao2024MiniCPM} & 2.8B & 71.9 & 55.6$^\mathsection$ & - & 74.1 & 605 \\ 
        InternVL 2~\cite{chen2024far}  & 2B & 86.9 & 76.2 & 58.9 & 73.4 & 784 \\
        \rowcolor[HTML]{F2F3F5} 
        Mini-Monkey (Ours) & 2B & \textbf{87.4} & \textbf{76.5} & \textbf{60.1} & \textbf{76.0}  & \textbf{806} \\ 
        \bottomrule
    \end{tabular}}}
    \vspace{-6pt}
\label{tab:vqa_metric}
\end{table*}

    \begin{table*}[t]
    \centering
    \caption{Quantitative accuracy (\%) comparison of our model with existing multimodal large language models (MLLMs) on several benchmarks. Following TextMonkey~\cite{liu2024textmonkey}, we use the accuracy metrics to evaluate our method.}
    \resizebox{1\linewidth}{!}{\begin{tabular}{c|cc|ccc|ccc}
    \toprule      \multirow{2}{*}{Method}
                 & \multicolumn{2}{c|}{Scene Text-Centric VQA}        & \multicolumn{3}{c|}{Document-Oriented VQA}                    & \multicolumn{3}{c}{KIE}   \\
                 & STVQA & TextVQA & DocVQA & InfoVQA & ChartQA & FUNSD   & SROIE  & POIE  \\ \midrule
    BLIP2-OPT-6.7B~\cite{li2023blip2}   & 20.9  & 23.5  & 3.2    & 11.3           & 3.4                  & 0.2     & 0.1    & 0.3  \\
    mPLUG-Owl~\cite{ye2023mplug}    & 30.5  & 34.0  & 7.4    & 20.0             & 7.9           & 0.5     & 1.7    & 2.5   \\
    InstructBLIP~\cite{dai2023instructblip} & 27.4  & 29.1   & 4.5    & 16.4           & 5.3             & 0.2     & 0.6    & 1.0     \\
    LLaVAR~\cite{zhang2023llavar}       & 39.2  & 41.8   & 12.3   & 16.5           & 12.2              & 0.5     & 5.2    & 5.9    \\
    BLIVA~\cite{hu2023bliva}        & 32.1  & 33.3  & 5.8    & 23.6           & 8.7             & 0.2     & 0.7    & 2.1     \\
    mPLUG-Owl2-8~\cite{ye2023mplugowl2}   & 49.8  & 53.9   & 17.9   & 18.9           & 19.4            & 1.4     & 3.2    & 9.9         \\
    LLaVA1.5-7B~\cite{liu2023llava1.5}     & 38.1  & 38.7    & 8.5    & 14.7           & 9.3            & 0.2     & 1.7    & 2.5          \\
    TGDoc~\cite{wang2023TGDoc}   & 36.3 & 46.2  & 9.0           & 12.8                & 12.7       & 1.4    & 3.0   & 22.2   \\
    UniDoc~\cite{feng2023unidoc}       & 35.2  & 46.2    & 7.7    & 14.7           & 10.9                    & 1.0       & 2.9    & 5.1    \\
    DocPedia~\cite{feng2023docpedia}     & 45.5  & 60.2   & 47.1   & 15.2           & 46.9                 & {29.9}    & 21.4   & 39.9    \\
    Monkey-8B~\cite{li2024monkey}       & {54.7}  & 64.3    & {50.1}   & {25.8}           & {54.0}            & 24.1    & {41.9}   & 19.9  \\ 
    InternVL-8B~\cite{chen2024internvl}     & 62.2  & 59.8      &28.7    & 23.6           & 45.6  & 6.5    & 26.4    & 25.9   \\
    InternLM-XComposer2-7B~\cite{dong2024internlmxcomposer2}      & 59.6 & 62.2      &39.7    & 28.6           & 51.6  & 15.3   & 34.2    & 49.3   \\ 
    TextMonkey-9B~\cite{liu2024textmonkey}   & 61.8 & 65.9    & 64.3          & 28.2                & 58.2       & 32.3    & 47.0   & 27.9   \\
    InternVL2-2B~\cite{chen2024far}   & 65.6 & 66.2    & 76.7          & 46.8                & \textbf{67.6}       & 42.0    & 68.0   & 66.8   \\
    \rowcolor[HTML]{F2F3F5} 
    Mini-Monkey-2B (Ours) & \textbf{67.2} & \textbf{68.8} & \textbf{78.4} & \textbf{50.0} & 67.3 & \textbf{43.2} & \textbf{70.5} & \textbf{71.2} \\ \bottomrule
    \end{tabular}}
    \label{tab:accuracy_metric}
    \end{table*}

\begin{table*}[!t]
    \small
    \caption{Ablation study of Complementary Image Pyramid. We compare our method with the existing cropping strategy and the overlay cropping strategy.}
    \centering
    \setlength{\tabcolsep}{1.5mm}{
    \resizebox{0.95\linewidth}{!}{\begin{tabular}{l|l|llllllll}
    \toprule
        Model & Resolution Strategy & TextVQA & OCRBench & MME & HallB & POPE \\ \midrule
        Baseline & Dynamic high-res Strategy~\cite{chen2024far} & 73.4 & 784 & 1876.8 & 37.9  & 85.2  \\ 
        Baseline & Fixed Size high-res Strategy~\cite{li2024monkey} & 74.2 & 772 & 1824.5 & 37.6  & 85.0  \\ 
        Baseline & Overlapping Cropping Strategy & 70.6 & 758 & 1874.1 & 36.8  & 83.5  \\ 
        Baseline & Multi-Scale Strategy~\cite{shi2024we} & 74.8 & 776 & 1846.8 & 38.1  & 85.3  \\ 
        Mini-Monkey (Ours) & Complementary Image Pyramid & \textbf{76.0} & \textbf{806} & \textbf{1884.2} & \textbf{38.8} & \textbf{88.0}  \\ 
        \bottomrule
    \end{tabular}}}
\label{tab:ab_CIP}
\end{table*}

\subsection{Comparison with State-of-the-art Methods}

\textbf{General Multimodal Understanding.} We evaluate Mini-Monkey on general multimodal understanding following ~\cite{he2024bunny,chen2024far}. The results are shown in Tab.~\ref{tab:generate_results}. Mini-Monkey surpasses other 2B-parameter models on 8 benchmarks. On the MathVista and POPE, Mini-Monkey outperforms the previous state-of-the-art method InternVL2-2B by 1\% and 2.8\%, respectively. On the HallusionBench, Mini-Monkey outperforms MiniCPM-V-2 by 2.7\%. These results showcase the ability of Mini-Monkey to handle general multimodal understanding and reasoning tasks.

\textbf{Document Understanding.} For the first type of metric, the results are presented in Tab.~\ref{tab:vqa_metric}. We use a relaxed accuracy measure for ChartQA, ANLS for DocVQA and InfoVQA, and the VQA score for TextVQA. The results indicate that Mini-Monkey achieves state-of-the-art performance among 2B-parameter multimodal large language models. Compared to InternVL2-2B, our method outperforms it by 2.6\%, 1.2\%, and 22 for TextVQA, InfoVQA, and OCRBench, respectively. Due to the small original resolution of ChartQA, it is less impacted by cropping operations, resulting in a minor improvement from our method. Notably, in the OCRBench, Mini-Monkey even surpasses the 8B-parameter Large Multimodal Model InternVL2-8B and the 9B-parameter Large Multimodal Model GLM4-V by 12 and 20, respectively. These results demonstrate the advantages of a Complementary Image Pyramid in enhancing document understanding. For the accuracy metric, the results are shown in Tab.~\ref{tab:accuracy_metric}. Mini-Monkey shows an average performance improvement of 15.6\% compared to TextMonkey-9B~\cite{liu2024textmonkey}, demonstrating the effectiveness of our method. Mini-Monkey also outperforms the state-of-the-art methods on multiple text-related benchmarks. Specifically, Mini-Monkey outperforms the InternVL2-2B by 2.6\%, 3.2\%, and 4.4\% on TextVQA, InfoVQA, and POIE, respectively. These results further indicate the great potential of Mini-Monkey for document understanding tasks.

\subsection{Ablation Study}
\label{sec:ab}
In this section, we perform ablation studies on both general multimodal understanding and document understanding benchmarks to validate the effectiveness of our method. We adopt the TextVQA~\cite{singh2019towards}, OCRBench~\cite{liu2023hidden}, HallusionBench~\cite{guan2023hallusionbench}, MME~\cite{fu2023mme}, and POPE~\cite{li2023evaluating} to conduct ablation studies.

\textbf{Complementary Image Pyramid.}
We conducted ablation studies to investigate the effectiveness of the CIP. We compared our method with several alternatives: The dynamic high-res strategy~\cite{chen2024far}, which maintains aspect ratios to increase resolution. The fixed-size high-res strategy~\cite{li2024monkey}, which uses a fixed size to increase resolution. The overlapping cropping strategy uses a high-res approach but crops with overlay. The multi-scale strategy~\cite{shi2024we}, which introduces a multi-scale strategy to the MLLM. As presented in Tab.~\ref{tab:ab_CIP}, the proposed CIP achieved the best results. Our method outperforms the previous multi-scale strategy~\cite{shi2024we} on both general multimodal understanding and document understanding. Notably, the over-overlay cropping strategy, instead of improving the model's performance, actually degraded it. 

\textbf{Various Model Capacity.} We performed ablation studies to assess the impact of CIP on models with varying model capacity. As illustrated in Table \ref{tab:ab_CIP_2}, CIP consistently improves the performance of varying model capacity. 

\textbf{Different Usage Configurations.} To further validate the improvements introduced by CIP, we performed experiments on usage configurations: a training-free configuration and a fine-tuning configuration. As shown in Tab.~\ref{tab:ab_CIP_3}, CIP demonstrates improvements in performance even when applied without training. The performance can be further improved with fine-tuning. Additionally, we surprisingly find that CIP can even facilitate the model fine-tuning process. As presented in the second line in Tab.~\ref{tab:ab_CIP_3}, direct fine-tuning of the Baseline model not only failed to improve performance but, in some cases, led to a decline. Conversely, incorporating CIP during the fine-tuning of the Baseline resulted in substantial improvements in both general multimodal understanding and document understanding, as evidenced in the fourth line of Tab.~\ref{tab:ab_CIP_3}. 

\textbf{Incorporating CIP to various MLLMs.} The proposed complementary image pyramid (CIP) can be seamlessly integrated into crop-based methods. To demonstrate its effectiveness, we incorporated CIP into various structures of MLLM, such as MiniCPM-V-2~\cite{yao2024MiniCPM}, InternVL 2~\cite{chen2024far}, LLaVA-OV~\cite{li2024llava}. As shown in Tab.~\ref{tab:ab_CIP_2}, CIP consistently enhances the performance across different MLLM structures, thereby validating the effectiveness of our approach. 

\textbf{The Number of Sub-Images.} To investigate whether the performance enhancement is attributed to an increase in the number of sub-images, we performed an experiment by incrementally raising the sub-image count for the Baseline. The findings, summarized in Table \ref{tab:ab_num_img}, indicate that increasing the number of sub-images does not lead to better performance; instead, it may result in a decline. In contrast, CIP can effectively improve the performance of the model. These results further demonstrate the effectiveness of the CIP.

\begin{table*}[!t]
    \small
    \caption{Exploring different usage configurations of CIP. Train represents fine-tuning the model.}
    \centering
    \setlength{\tabcolsep}{1.5mm}
    \resizebox{0.75\linewidth}{!}{\begin{tabular}{l|l|l|lllllll}
    \toprule
        Model & CIP & Train & TextVQA & OCRBench & MME & HallB & POPE \\ \midrule
        InternVL2-2B & $\times$ & $\times$ & 73.4 & 784 & 1876.8 & 37.9  & 85.2  \\
        InternVL2-2B & $\times$ & $\checkmark$ & 73.3 \textcolor{red}{\textbf{(-0.1)}} & 787 \textcolor{red}{\textbf{(+3)}} & 1858.3 \textcolor{red}{\textbf{(-18.5)}} & 37.3  \textcolor{red}{\textbf{(-0.6)}} & 85.3 \textcolor{red}{\textbf{(+0.1)}} \\
        InternVL2-2B & $\checkmark$ & $\times$ & 75.2 \textcolor{red}{\textbf{(+1.8)}} & 800 \textcolor{red}{\textbf{(+17)}} & 1881.9 \textcolor{red}{\textbf{(+5.1)}} & 38.7 \textcolor{red}{\textbf{(+0.8)}} & 86.7 \textcolor{red}{\textbf{(+1.5)}} \\
        InternVL2-2B & $\checkmark$ & $\checkmark$ & 76.0 \textcolor{red}{\textbf{(+2.6)}} & 806 \textcolor{red}{\textbf{(+22)}} & 1884.2 \textcolor{red}{\textbf{(+7.4)}} & 38.8 \textcolor{red}{\textbf{(+0.9)}} & 88.0 \textcolor{red}{\textbf{(+2.8)}} \\
    \bottomrule
    \end{tabular}}
    \label{tab:ab_CIP_3}
\end{table*}

\begin{table*}[!t]
    \small
    \caption{Ablation study of incorporating complementary image pyramid (CIP) to other MLLMs. \textsuperscript{$\mathsection$} represents the results from the OpenCompass leaderboard~\cite{2023opencompass}.}
    \centering
    \setlength{\tabcolsep}{1.5mm}
    \resizebox{0.75\linewidth}{!}{\begin{tabular}{l|l|llllllll}
    \toprule
        Model & CIP & TextVQA & OCRBench & MME & HallB & POPE \\ \midrule
        MiniCPM-V-2-2.8B & $\times$ & 74.1 & 605 & 1808.6 & 36.1\textsuperscript{$\mathsection$}  & 86.3\textsuperscript{$\mathsection$}  \\
        MiniCPM-V-2-2.8B & $\checkmark$ & 76.0 \textcolor{red}{\textbf{(+1.9)}} & 627 \textcolor{red}{\textbf{(+22)}} & 1819.5 \textcolor{red}{\textbf{(+10.9)}} & 36.5 \textcolor{red}{\textbf{(+0.4)}}  & 87.1 \textcolor{red}{\textbf{(+0.8)}}  \\
        LLaVA-OV-0.5B & $\times$ & 65.3 & 577 & 1478.0 & 28.1  & 86.7  \\
        LLaVA-OV-0.5B& $\checkmark$ & 66.2 \textcolor{red}{\textbf{(+0.9)}} & 600 \textcolor{red}{\textbf{(+23)}} & 1482.6 \textcolor{red}{\textbf{(+4.6)}} & 28.8 \textcolor{red}{\textbf{(+0.7)}} & 87.7 \textcolor{red}{\textbf{(+1.0)}} \\
        InternVL2-1B & $\times$ & 70.5 & 754 & 1794.4 & 33.4 & 84.9\textsuperscript{$\mathsection$}  \\
        InternVL2-1B & $\checkmark$ & 72.3 \textcolor{red}{\textbf{(+1.8)}} & 772 \textcolor{red}{\textbf{(+18)}} & 1801.5 \textcolor{red}{\textbf{(+7.1)}} & 34.3 \textcolor{red}{\textbf{(+0.9)}} & 85.7 \textcolor{red}{\textbf{(+0.8)}} \\
        InternVL2-2B & $\times$ & 73.4 & 784 & 1876.8 & 37.9  & 85.2  \\
        InternVL2-2B & $\checkmark$ & 76.0 \textcolor{red}{\textbf{(+2.6)}} & 806 \textcolor{red}{\textbf{(+22)}} & 1884.2 \textcolor{red}{\textbf{(+7.4)}} & 38.8 \textcolor{red}{\textbf{(+0.9)}} & 88.0 \textcolor{red}{\textbf{(+2.8)}} \\
        InternVL2-8B & $\times$ & 77.4 & 794 & 2210.3 & 45.0\textsuperscript{$\mathsection$} & 84.2\textsuperscript{$\mathsection$}  \\
        InternVL2-8B & $\checkmark$ & 79.3 \textcolor{red}{\textbf{(+1.9)}} & 809 \textcolor{red}{\textbf{(+15)}} & 2226.4 \textcolor{red}{\textbf{(+16.1)}} & 45.4 \textcolor{red}{\textbf{(+0.4)}} & 84.8 \textcolor{red}{\textbf{(+0.6)}} \\
    \bottomrule
    \end{tabular}}
    \label{tab:ab_CIP_2}
\end{table*}

\begin{table*}[!t]
    \small
    \caption{Ablation study of the scale compression mechanism. We used different compression ratios to compare with FastV~\cite{chen2024image}. (0.5) represents 50\% compression and (0.9) represents 90\% compression.}
    \centering
    \setlength{\tabcolsep}{1.5mm}{
    \resizebox{0.7\linewidth}{!}{\begin{tabular}{l|l|llllllll}
    \toprule
        Model & Resolution Strategy & TextVQA & OCRBench & MME & HallB & POPE \\ \midrule
        Mini-Monkey & Pooling (0.5) & 47.6 & 256 & 1765.2 & 31.5 & 84.5 \\ 
        Mini-Monkey & Random (0.5) & 63.5 & 503 & 1805.5 & 36.2 & 85.9 \\ 
        Mini-Monkey & FastV~\cite{chen2024image} (0.5) & 73.4 & 781 & 1848.0 & 38.3 & 83.9 \\ 
        Mini-Monkey & FastV~\cite{chen2024image} (0.9) & 73.9 & 792 & 1866.1 & 37.5 & 85.8 \\ 
        Mini-Monkey & SCM (0.5) & 74.7 & 794 & \textbf{1886.0} & \textbf{38.7} & 86.1 \\ 
        Mini-Monkey & SCM (0.9) & \textbf{75.2} & \textbf{801} & 1884.7 & 38.6 & \textbf{86.2} \\ 
        \bottomrule
    \end{tabular}}}
\label{tab:ab_compress}
\end{table*}

\textbf{Scale Compression Mechanism.} We compared the proposed Scale Compression Mechanism with the related work FastV~\cite{chen2024image}. For different methods, we compress the number of visual tokens by 50\%. For our method and FastV, we further conduct an experiment with 90\% compression. Following FastV's paper, we set the K in FastV as 2. As illustrated in Tab.~\ref{tab:ab_compress}, when using 50\% compression and 90\% compression, our method outperformed FastV by 21.5\% and 4.4\%, respectively, demonstrating its effectiveness. FastV compresses input tokens, including both visual and textual tokens, within Transformer blocks. In contrast, our method works in conjunction with the CIP and more targeted by using tokens with high relative information density to compress tokens with low information density.

\begin{table*}[!t]
    \small
    \caption{Ablation study on the number of sub-images. The number denotes the sub-image count.}
    \centering
    \setlength{\tabcolsep}{1.5mm}{
    \resizebox{0.55\linewidth}{!}{\begin{tabular}{l|l|llllllll}
    \toprule
        Model & Number & TextVQA & OCRBench & MME & HallB & POPE \\ \midrule
        Baseline & 18 & 74.2 & 782 & 1851.7 & 37.0 & 85.8 \\ 
        Baseline & 24 & 74.4 & 783 & 1857.6 & 36.9 & 85.8 \\ 
        Baseline & 32 & 74.3 & 782 & 1845.0 & 36.9 & 85.9 \\ 
        Baseline & 48 & 74.0 & 767 & 1841.6 & 36.2 & 85.7 \\ 
        CIP & 32 & \textbf{76.0} & \textbf{806} & \textbf{1884.2} & \textbf{38.8} & \textbf{88.0} \\ 
        \bottomrule
    \end{tabular}}}
\label{tab:ab_num_img}
\end{table*}

\begin{figure*}[!t]
    \centering
    \includegraphics[width=\linewidth]{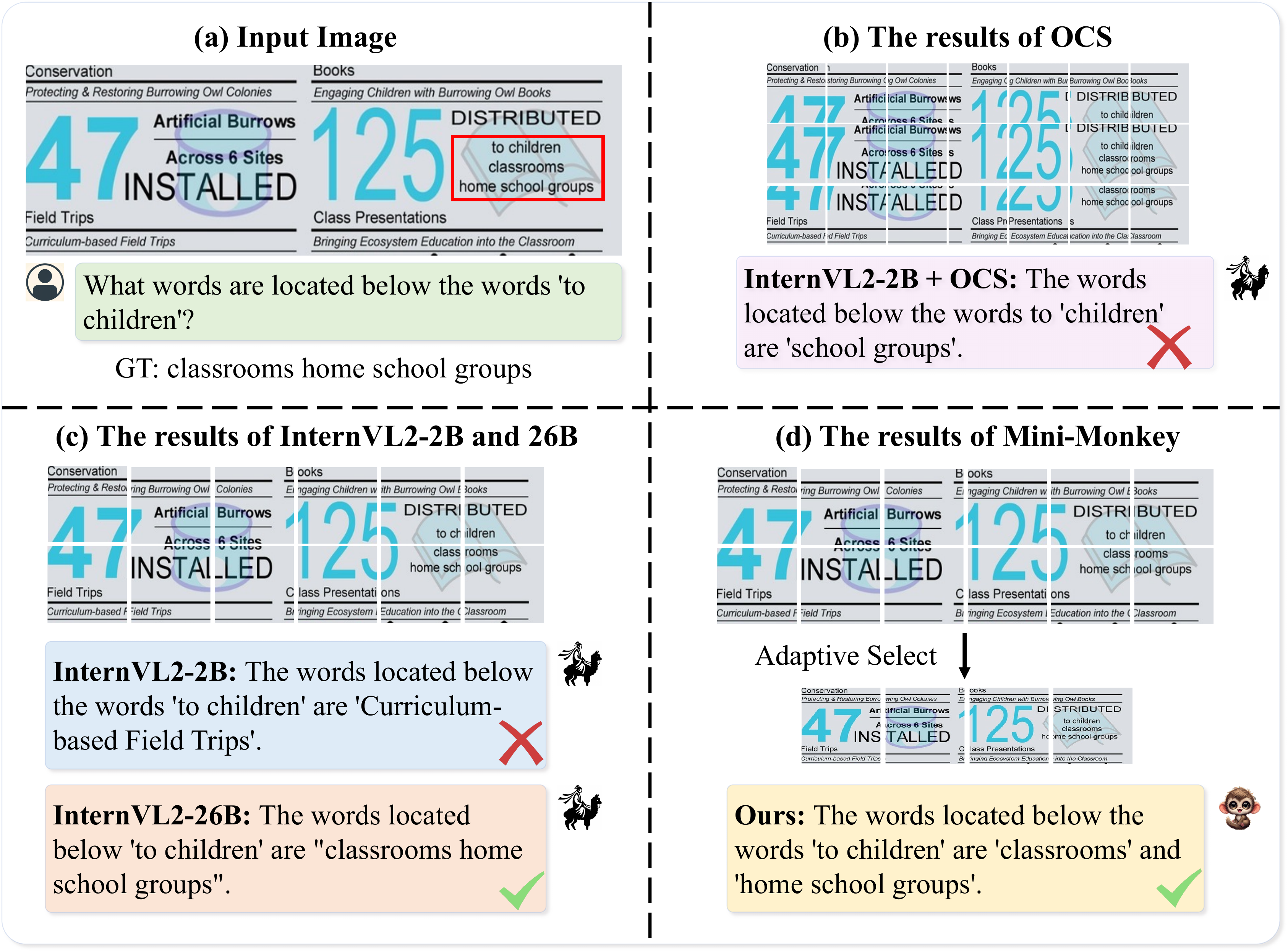}
    \caption{Qualitative results of Mini-Monkey. (a) Input Image and Ground Truth. (b) The results of using overlapping cropping strategy. OSC represents the overlapping cropping strategy. (c) The results of InternVL2-2B and InternVL2-26B. (d) The results of Mini-Monkey.}
    \label{fig:visualization}
\end{figure*}

\subsection{Qualitative Results}
\label{qualitative}
In this section, we provide some qualitative results to demonstrate the effectiveness of our method. First, we verify that the semantic sawtooth effect is particularly evident in lightweight MLLMs, which adopt InternVL2-2B and InternVL2-26B. As shown in Fig.~\ref{fig:visualization}(c), InternVL2-26B can answer the questions correctly. However, due to the word `classrooms' and `school' being cropped, InternVL2-2B gives a wrong answer that addresses the text in the bottom left corner of the original image. Mini-Monkey can overcome this semantic sawtooth effect and provide the correct answer, as presented in Fig.~\ref{fig:visualization}(d). Comparing Fig.~\ref{fig:visualization}(b) and Fig.~\ref{fig:visualization}(d), we can see that the overlapping cropping strategy introduces some hallucinations and cannot answer questions accurately based on the image, whereas our methods can effectively address the semantic sawtooth effect.

\section{Conclusion}    
In this study, we introduce a Complementary Image Pyramid (CIP) designed to alleviate the semantic sawtooth effect for MLLMs, thereby enhancing their capability to process high-resolution images effectively. CIP is plug-and-play and can be seamlessly integrated into various multimodal large language models at a low cost. We demonstrate the effectiveness of the proposed CIP across diverse architectures, various parameters, and different usage configurations, leading to consistent performance improvements. Besides, we present a Scale Compression Mechanism (SCM) to compress the visual tokens for computational efficiency. CIP not only enhances the general multimodal understanding performance but also shows consistent improvements in document understanding tasks. Furthermore, our experimental results demonstrate that 2B-parameter MLLM equipped with CIP even surpasses larger 8B-parameter state-of-the-art models like InternVL2-8B on the OCRBench. \textbf{Limitations:} To ensure the seamless application of our method across various architectures, we adopt an image-centric approach to construct an image pyramid without introducing additional parameters. In future work, we will explore the use of trainable Feature Pyramid Network (FPN) for MLLMs, aiming to more efficiently leverage multi-scale features.

{
\bibliographystyle{plain}
 \bibliography{arxiv_clean}

}

\end{document}